\ifdefined\pdfminorversion
\fi
\documentclass{ijSmartGrid}

\newcommand{\venuefooter}{\footnotesize 40th Conference on Neural Information Processing Systems (NeurIPS 2026). Track on Evaluations and Datasets. (Oral)}
\makeatletter
\fancypagestyle{plain}{%
  \fancyhf{}%
  \fancyhead[L]{\@ijfullheader}%
  \fancyfoot[L]{\venuefooter}%
}
\makeatother

\usepackage{multirow}
\usepackage{xcolor}
\usepackage{fontawesome5}
\definecolor{forestgreen}{rgb}{0.0, 0.5, 0.0}
\definecolor{ashgrey}{rgb}{0.7, 0.75, 0.71}
\definecolor{darkorange}{rgb}{1.0, 0.55, 0.0}
\usepackage{pifont}
\usepackage{makecell}
\usepackage{enumitem}
\definecolor{generalcolor}{HTML}{0185F9}
\definecolor{rscolor}{HTML}{fc8e62}
\definecolor{ftcolor}{HTML}{95EC69}
\definecolor{ques}{RGB}{192,0,0}
\definecolor{revision}{RGB}{255,0,0}
\newcommand{\generalcolor}[1]{\textcolor{generalcolor}{#1}}
\newcommand{\rscolor}[1]{\textcolor{rscolor}{#1}}
\newcommand{\ftcolor}[1]{\textcolor{ftcolor}{#1}}
\newcommand{\gb}{\generalcolor{$\bullet$\,}}
\newcommand{\rsb}{\rscolor{$\bullet$\,}}
\newcommand{\ftb}{\ftcolor{$\bullet$\,}}

\usepackage{tcolorbox}
\tcbuselibrary{listings, breakable, skins}
\definecolor{utblue}{RGB}{0,114,178}

\newtcolorbox{promptbox}[1][]{%
  enhanced,
  breakable,
  colback=utblue!4,
  colframe=utblue!60,
  boxrule=0.5pt,
  arc=3pt,
  left=8pt, right=8pt, top=6pt, bottom=6pt,
  fontupper=\footnotesize\ttfamily,
  before upper={\setlength{\parskip}{4pt}\setlength{\parindent}{0pt}},
  title=#1,
  fonttitle=\bfseries\sffamily,
  colbacktitle=utblue!60,
  coltitle=white,
  attach boxed title to top left={xshift=8pt, yshift=-3pt},
  boxed title style={colframe=utblue!60, sharp corners, arc=0pt},
}

\title{Can LLM Agents Respond to Disasters? Benchmarking Heterogeneous Geospatial Reasoning in Emergency Operations}

\author{%
  Junjue Wang$^{1,2*}$, Weihao Xuan$^{1,2*}$, Heli Qi$^{2}$, Pengyu Dai$^{1,2}$, Kunyi Liu$^2$, Hongruixuan Chen$^2$, \\ Zhuo Zheng$^3$, Junshi Xia$^2$, Stefano Ermon$^3$, Naoto Yokoya$^{1,2\dagger}$ \\
}

\ijaffiliation{%
  \textsuperscript{1}The University of Tokyo, 
  \textsuperscript{2}RIKEN AIP, 
  \textsuperscript{3}Stanford University\\
  \textsuperscript{*}Equal Contribution
  \textsuperscript{\dag}Corresponding Author
}

\ijshortauthor{J.~Wang et al.}

\ijvolume{x}
\ijissue{x}
\ijpubmonth{Month}
\ijpubyear{xxxx}

\begin{document}

\maketitle

\begin{abstract}
Operational disaster response goes beyond damage assessment, requiring responders to integrate multi-sensor signals, reason over road networks, populations and key facilities, plan evacuations, and produce actionable reports. However, prior work largely isolates remote-sensing perception or evaluates generic tool use, leaving the end-to-end workflows of emergency operations underexplored. In this paper, we introduce \textbf{D}isaster \textbf{O}perational \textbf{R}esponse \textbf{A}gent benchmark (\textbf{DORA}), the first agentic benchmark for end-to-end disaster response: 515 expert-authored tasks across 45 real-world disaster events spanning 10 types, paired with expert-verified, replayable gold trajectories totaling 3{,}500 tool-call steps. Tasks span five dimensions that cover the operational disaster-response pipeline: disaster perception, spatial relational analysis, disaster operational planning, temporal evolution reasoning, and multi-modal report synthesis. Agents compose calls from a 108-tool MCP library over heterogeneous geospatial data: optical, SAR, and multi-spectral imagery across single-, bi-, and multi-temporal sequences (0.015--10\,m GSD), complemented by elevation and social vector layers. 
We comprehensively 
evaluate 13 frontier LLMs on our benchmark, revealing three persistent challenges: 1)~disaster-domain grounding exposes unique failure modes (damage-semantic grounding, sensor-modality mismatch, and disaster-pipeline composition); 2)~agents are doubly bottlenecked by tool selection and argument grounding, where gold tool-order hints improve accuracy by only 1.08--4.40\%, and alternative scaffolds yield at most a 3.24\% gain; 3)~compositional fragility scales with trajectory length, the agent-to-gold gap widening from 7\% to 56\% on long pipelines.
DORA establishes a rigorous testbed for operationally reliable disaster-response agents. Code and data are available at \url{https://github.com/Junjue-Wang/DORA}.
\end{abstract}


\section{Introduction}
Natural and man-made disasters (earthquakes, floods, hurricanes, landslides,
and explosions) claim tens of thousands of lives annually and inflict
hundreds of billions of dollars in infrastructure damage~\cite{xu2025implementing,frankenberg2020effects}.
Effective disaster response requires compound analytical capabilities:
perceiving damage from heterogeneous sensor data, reasoning
about spatial relationships among affected assets, estimating 
operational resources under real-world constraints, tracking 
disaster evolution over time, and synthesizing findings 
into actionable reports.
These tasks cannot be addressed by visual inspection alone but demand the tight integration of multi-sensor remote sensing (RS) imagery, social geospatial vector data, domain-specific analytical tools, and multi-step compositional reasoning.

Recent advances in LLM-based agents have demonstrated strong capabilities in multi-step tool use, compositional reasoning, and long-horizon task execution across general-purpose domains (web browsing, code generation, database querying, etc.)~\cite{wang2024executable,yang2024swe,liu2024agentbench}.
These developments have also begun to push the boundaries of remote sensing applications: agents can now automate geospatial workflows~\cite{feng2026earthagent, shabbir2026openearthagent,zhao2026openearth}, answer multi-step queries over satellite imagery with specialized tools.
However, disaster operations uniquely combine heterogeneous data fusion, long compositional pipelines, and disaster-specific knowledge grounding that prior agent and RS benchmarks rarely test together, leaving the full operational pipeline of disaster response largely unexplored.
To this end, we introduce DORA, a disaster agent benchmark grounded in 45 real-world disaster events that systematically 
evaluates state-of-the-art LLM agents across diverse disaster scenarios, providing a reliable AI platform for advancing humanitarian efforts in emergency response.

\begin{figure*}[hbt]
    \centering
    \includegraphics[width=1\linewidth]{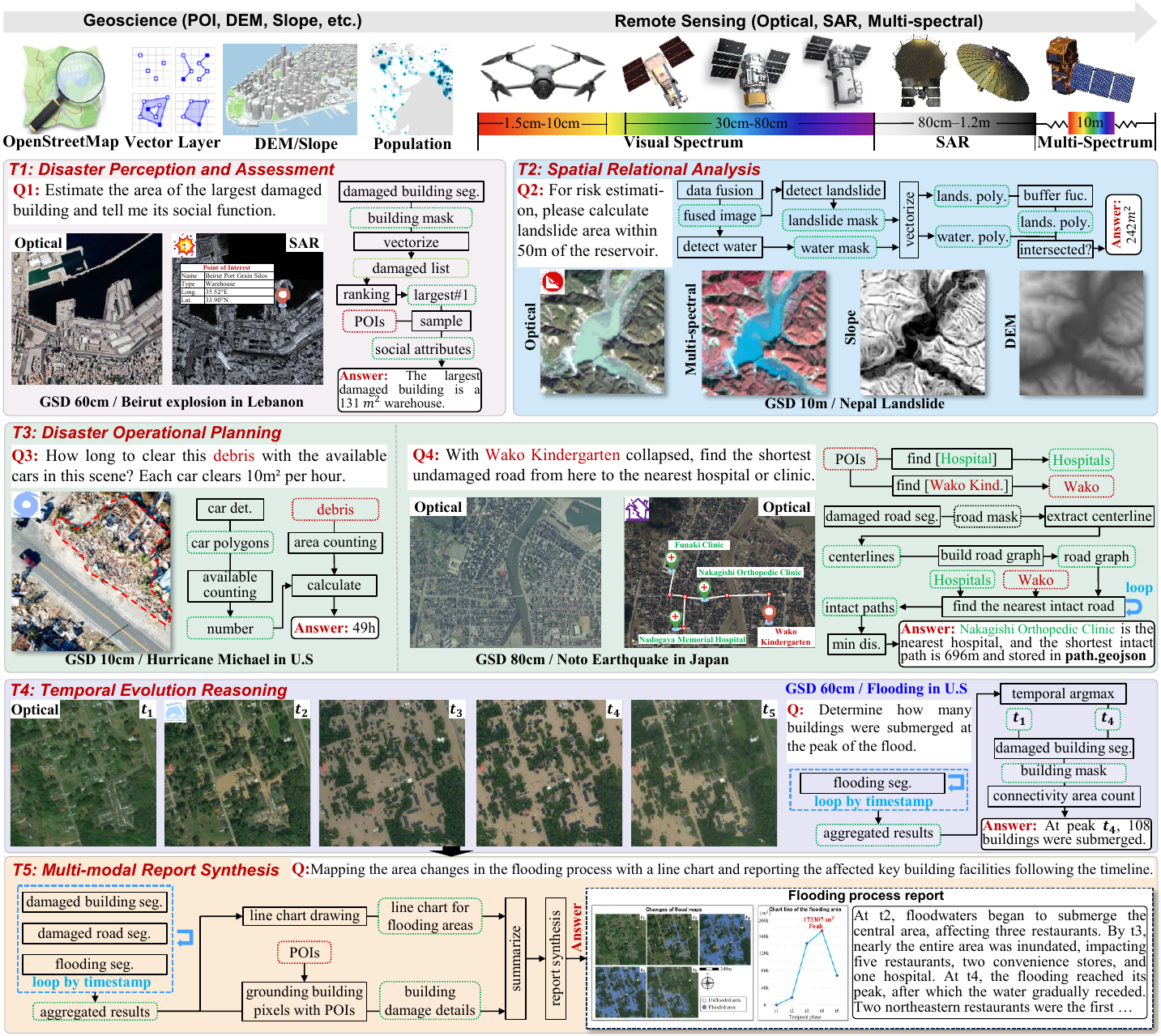}
    \caption{Representative task examples in DORA across different disaster operational categories. Each example illustrates the heterogeneous data inputs and multi-step tool chains required to produce concrete, decision-ready outputs.} 
\label{fig:dataset_vis}
\end{figure*}

To comprehensively evaluate LLM agents with diverse disaster-response capabilities, we organize tasks \textbf{into} five complementary analytical dimensions that collectively span the information-processing demands of disaster intelligence: 1) disaster perception and assessment, 2) spatial relational analysis, 3) disaster operational planning, 4) temporal evolution reasoning, and 5) multi-modal report synthesis.
A single disaster scenario may simultaneously require capabilities from multiple dimensions, and each dimension exercises a distinct combination of tools, reasoning patterns, and data modalities. This design ensures that DORA measures not merely whether an agent can invoke individual tools, but whether it can \textbf{translate} operational knowledge into executable tool pipelines: selecting the right tools, composing them in the correct order, and interpreting intermediate outputs to produce decision-ready results.
Our main contributions are:

\begin{enumerate}[leftmargin=*,itemsep=2pt]
  \item We introduce DORA, the first agentic benchmark for operational disaster response, comprising 515 expert-authored tasks grounded in 45 real-world disaster sites spanning 10 disaster types (hurricanes, earthquakes, floods, wildfires, etc.), with expert-verified gold trajectories totaling 3,500 tool-call steps across five complementary analytical dimensions from atomic perception to multi-modal report synthesis.
  
  \item We design and release a purpose-built geospatial MCP tool library of 108 tools organized into six functional modules (perception, raster, vector, logic, visualization, summarization) covering RS interpretation, spatial computation, routing, POI querying, logistics estimation, and report generation, the most comprehensive disaster-specific tool ecosystem to date.
  
  \item We benchmark 13 frontier LLMs through dimension-wise failure-mode, instruction-following, modality-stratified, trajectory-length, and scaffolding analyses, surfacing three persistent challenges: disaster-domain grounding exposes unique failure modes; agents are doubly bottlenecked by tool selection and argument usage (neither oracle tool-order hints nor alternative scaffolds close the gap); and compositional fragility scales sharply with trajectory length, revealing concrete directions for future disaster-response agent design.
\end{enumerate}
\section{Related Work}
\label{gen_inst}
\noindent \textbf{General LLM Agents.}
A key challenge in building LLM agents is closing the loop between reasoning and execution, motivating work on planning, self-correction, and experiential learning.
ReAct~\cite{yao2023react} establishes the core paradigm of interleaving reasoning traces with environment actions, enabling dynamic planning within a single inference loop.
ExpeL~\cite{zhao2024expel} extends this with experiential learning by extracting insights from accumulated trajectories, and AutoGuide~\cite{fu2024autoguide} generates context-aware guidelines from offline experiences to steer agents in unfamiliar domains.
More recently, ReasoningBank~\cite{ouyang2025reasoningbank} distills generalizable reasoning strategies from self-judged outcomes, enabling agents to self-evolve across task streams.
To systematically measure these capabilities, WebArena~\cite{zhou2024webarena}, OSWorld~\cite{xie2024osworld}, and AgentBench~\cite{liu2024agentbench} benchmark web navigation, desktop control, and multi-environment digital tasks, while SWE-bench~\cite{yang2025swebench}, OctoBench~\cite{ding2026octobench}, and FeatureBench~\cite{zhou2026acebench} target software-engineering workflows.
GAIA~\cite{mialon2024gaia} further probes general assistants on multi-step tool-use tasks. Unlike these benchmarks operating in digital environments, DORA evaluates LLM agents on real-world disaster response, where heterogeneous geospatial data, domain-specific tool orchestration, and operational decision-making collide in a single task.

\begin{table*}[!hbt]
\centering
\caption{Comparison of DORA with representative agent benchmarks.}
\label{tab:comparison}
\small
\setlength{\tabcolsep}{3.5pt}
\resizebox{1.0\linewidth}{!}{
\begin{tabular}{l l c c c c c c c c}
\toprule
\textbf{Benchmark}
  & \textbf{Domain}
  & \textbf{\#Tasks}
  & \textbf{\#Tools}
  & \textbf{\#Modality}
  & \textbf{GSD(m)}
  & \textbf{Annotate}
  & \textbf{Temporal}
  & \textbf{Eval Level}
  & \textbf{Avg Steps} \\
\midrule
GAIA~\cite{mialon2024gaia}
  & General QA     & 466    & --   & 2 & --         & Human   & --          & Final   & -- \\
GTA~\cite{wang2024gta}
  & General tools  & 229    & 14 & 2 & --         & Human    & --          & Step+Final     & 2.4 \\
m\&m's~\cite{ma2024m}
  & Multi-modal    & 4K+    & 33   & 2 & --         & Auto     & --          & Step+Final     & 2.7 \\
WebArena~\cite{zhou2024webarena}
  & Web navigation & 812    & -- & 2 & --         & Human    & --          & Execution     & -- \\
AgentBench~\cite{liu2024agentbench}
  & Digital envs   & 8 envs & -- & 2 & --         & Human    & --          & Execution      & -- \\
\midrule
GeoLLM-Engine~\cite{singh2024geollm}
  & Geospatial     & 500K+  & 175+ & 1 & 0.5--30   & Auto     & 1   & Final    & 5.2 \\
ThinkGeo~\cite{shabbir2025thinkgeo}  & Remote sensing & 486    & 14   & 2 & 0.1--30   & Human   & 1+2   & Step+Final     & 3.6 \\
UniVEarth~\cite{kao2025towards} & Remote sensing & 140    & --   & 1 & 15-1000       & Human   & 1      & Final   & -- \\
Earth-Agent~\cite{feng2026earthagent} & Remote sensing & 248    & 104  & 3 & 0.3--10   & Human   & 1+N & Step+Final    & 5.4 \\
OpenEarthAgent~\cite{shabbir2026openearthagent}
  & Remote sensing & 1169   & 28 & 4 & 0.1--30   & LLM+Valid. & 1+2 &  Step+Final & 6.0 \\
\midrule
\textbf{DORA (Ours)}
  & \textbf{Disaster ops.}
  & \textbf{515}
  & \textbf{108}
  & \textbf{8}
  & \textbf{0.015--10}
  & \textbf{Human}
  & \textbf{1+2+N}
  & \textbf{Step+Final}
  & \textbf{6.8} \\
\bottomrule
\end{tabular}}
\end{table*}

\noindent \textbf{Earth observation LLM Agents.}
Early RS agents treat LLMs as orchestrators of specialized visual models: RS-Agent~\cite{xu2024rs} invokes RS models for multi-step interpretation, and Change-Agent~\cite{liu2024change} focuses on bi-temporal change analysis with LLM-driven captioning.
GeoLLM-Engine~\cite{singh2024geollm} offers a realistic copilot environment reflecting analyst workflows, and ThinkGeo~\cite{shabbir2025thinkgeo} introduces a 486-task RS benchmark under a ReAct-style loop.
UniVEarth~\cite{kao2025towards} grounds queries in Google Earth Engine API calls, while Earth-Agent~\cite{feng2026earthagent} and OpenEarthAgent~\cite{shabbir2026openearthagent} couple multi-step reasoning with executable tools.
However, these rely on general geospatial datasets, LLM-synthesized queries, and general-purpose toolsets, missing the operational complexity of disaster response.
Disaster-specific \emph{perception} resources like xBD~\cite{gupta2019xbd} and DisasterM3~\cite{wang2025disasterm3} supply expert-verified damage masks and single-step VQA benchmarks.
DORA reuses their imagery but advances evaluation from perception and short-form QA to end-to-end operational reasoning: agents \emph{compose} heterogeneous tools across perception, routing, logistics, and synthesis to produce operationally actionable outputs (rescue routes, resource allocations, multi-modal briefings) beyond any single perception model.
DORA exposes three disaster-critical competencies untested elsewhere (Fig.~\ref{fig:failure_mode}): 1) \emph{cross-modal reasoning}, i.e., choosing optical, SAR, vector, or DEM under cloud, night, or terrain constraints, appears in 28.7\% of DORA tasks versus $<$3\% in prior RS benchmarks.
2) \emph{damage-semantic grounding} (e.g., flood vs.\ landslide debris, collapsed vs.\ flooded buildings) is the dominant failure mode (20.3\%) for frontier LLMs.
3) \emph{disaster pipeline composition} accounts for 56.3\% of errors, which stem from pipeline structure rather than individual tool misuse.

\section{DORA Dataset}
\label{sec:dataset}
\subsection{Data source overview}
\begin{figure*}[hbt]
    \centering
    \includegraphics[width=1\linewidth]{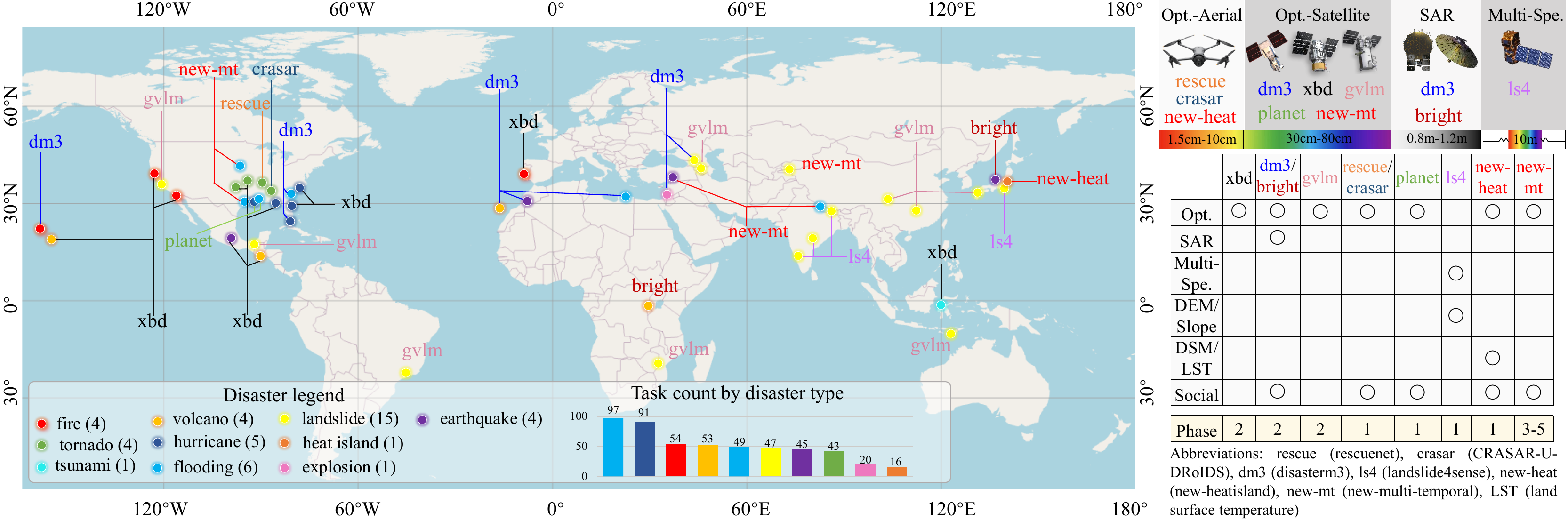}
    \caption{DORA aggregates multi-modal data from 10 open-source databases into 45 disaster events distributed across five continents, including 2,850 remote sensing images, 460 vector layers and 515 expert-designed tasks across 10 disaster types. `new' denotes our self-constructed samples.} 
\label{fig:distribution}
\end{figure*}

As shown in Fig.~\ref{fig:distribution}, 
we broadly collect open-source disaster imagery and geospatial resources to ensure diversity in disaster types, sensor modalities, and geographic coverage.
High-resolution optical pre- and post-disaster satellite image pairs are sourced from the xBD dataset~\cite{gupta2019xbd} (covering hurricanes, earthquakes, wildfires, volcanic eruptions, floods, tsunamis, and tornadoes), complemented by optical--SAR pairs from DisasterM3~\cite{wang2025disasterm3} and BRIGHT~\cite{chen2025bright}.
For multi-temporal sequences, we newly collect flood progression and post-disaster reconstruction scenes spanning 3--5 observation phases from NAIP~\cite{naip} and the Maxar Open Data Program~\cite{maxar_open_data}.
For multi-spectral terrain scenes, Landslide4Sense~\cite{ghorbanzadeh2022outcome} and GVLM-CD~\cite{ZHANG20231} provide composites with DEM and slope layers, further complemented by Planet imagery for urban flood risk analysis.
For aerial imagery, we incorporate RescueNet~\cite{rahnemoonfar2023rescuenet} and CRASAR-U-DRoIDS~\cite{manzini2024crasar} for fine-grained disaster scene analysis.
For heat-island analysis, orthophotos and digital surface and elevation models come from the Tokyo Metropolitan Government Digital Twin point cloud data~\cite{tmg_digitaltwin}, and mean radiant temperature maps are computed from them with SOLWEIG~\cite{lindberg2008solweig} and meteorological observations of the Japan Meteorological Agency~\cite{jma}.
Co-registered social geospatial layers come from OpenStreetMap~\cite{openstreetmap} and WorldPop~\cite{worldpop}, including POIs, road networks, population rasters, and facility footprints.

\subsection{Agent tasks in the context of disasters}
\label{sec3:tasks}
\begin{figure}[!hbt]
    \centering
    \begin{minipage}[b]{0.58\textwidth}
        \centering
        \includegraphics[width=\textwidth]{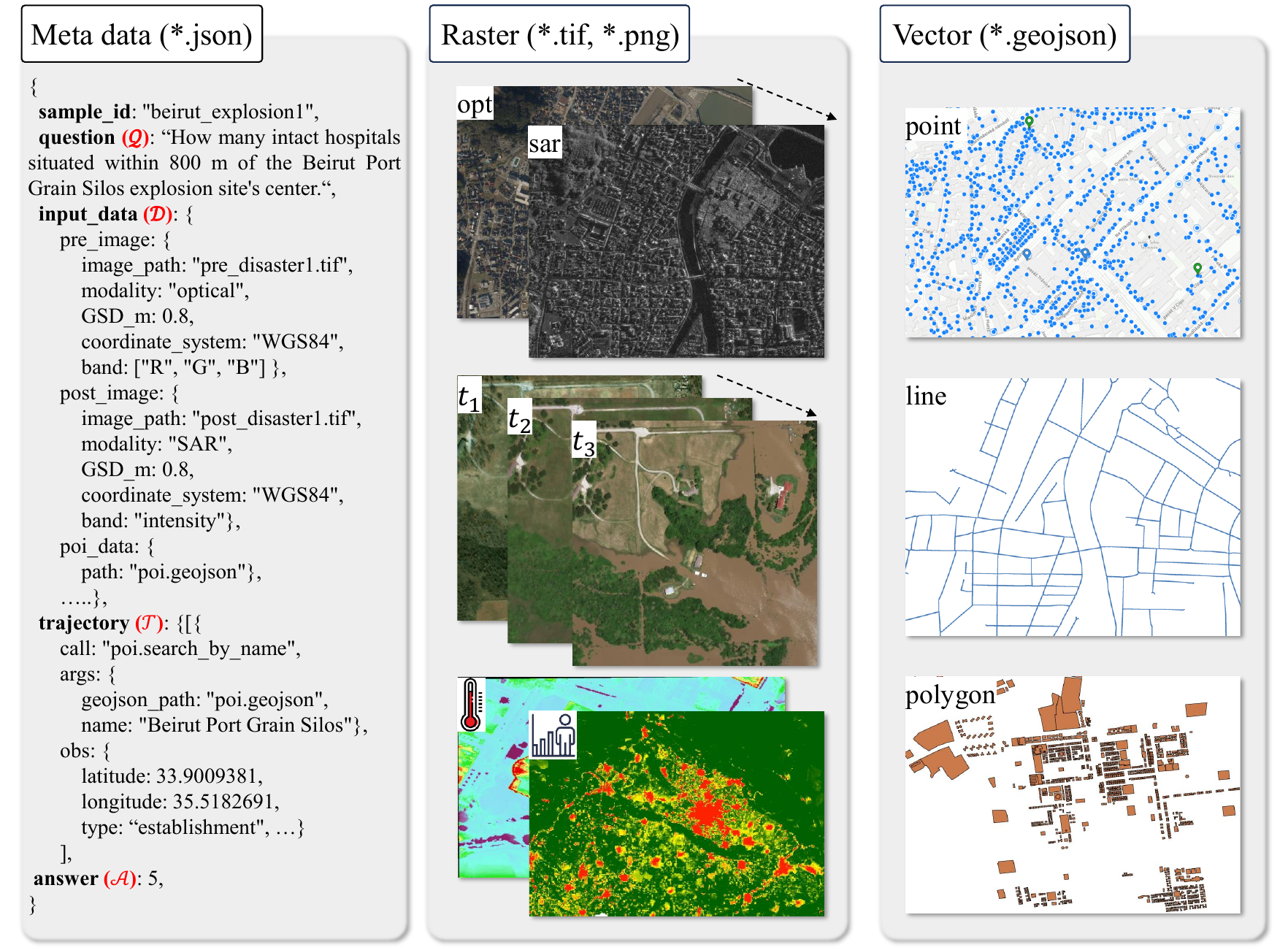}
        \caption{Each sample is stored as a JSON meta file linking the query ($\mathcal{Q}$), heterogeneous input data ($\mathcal{D}$, rasters and vectors), tool-call trajectory ($\mathcal{T}$), and final answer ($\mathcal{A}$).}
        \label{fig:heterogeneous_data}
    \end{minipage}
    \hfill
    \begin{minipage}[b]{0.40\textwidth}
        \centering
        \includegraphics[width=\textwidth]{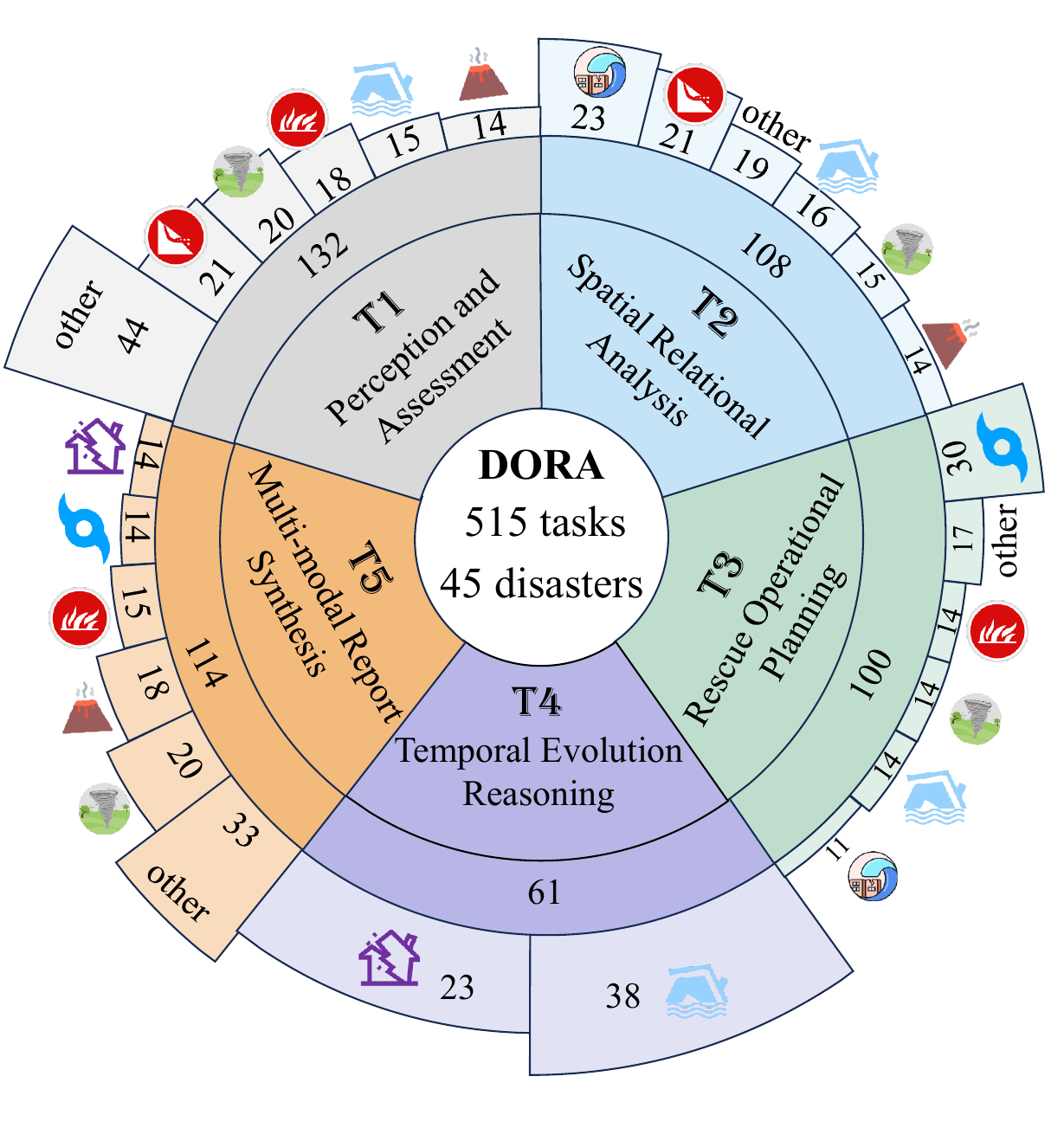}
        \caption{Distribution of 515 tasks across five analytical dimensions (inner ring) and disaster types (outer bars).}
        \label{fig:analytical_dim}
    \end{minipage}
\end{figure}

\noindent \textbf{Task formulation.} 
As shown in Fig.~\ref{fig:heterogeneous_data}, each DORA task is defined as a tuple $(\mathcal{Q},\mathcal{D},\mathcal{T}^{*},\mathcal{A}^{*})$. $\mathcal{Q}$ is a natural-language query grounded in an operational need. $\mathcal{D}$ is a \emph{heterogeneous data manifest} that bundles georeferenced raster layers across optical, SAR, and multi-spectral modalities (with metadata on modality, pixel size, coordinate reference system, band specification, etc.) alongside social geospatial vector layers (POIs, road networks, facility footprints) in GeoJSON format. $\mathcal{T}^{*}=\langle t_1,\dots,t_K\rangle$ is the expert-verified gold tool-call trajectory, and $\mathcal{A}^{*}$ is a structured JSON final answer whose fields (apart from rendered visualizations) fall into seven typed categories: \emph{scalar} (count, area, ratio), \emph{string} (damage grade, disaster type), \emph{point} (epicenter, target location), \emph{line} (shortest-path route), \emph{polygon} (damage extent, flood boundary), \emph{set} (affected facilities, buildings), and \emph{dict} (per-phase statistics, composite situation reports). At inference time the agent observes only $(\mathcal{Q},\mathcal{D})$ and must autonomously compose the correct tool chain from the 108-tool library (\S\ref{sec:tools}) to produce $\mathcal{A}$.

\begin{figure*}[hbt]
    \centering
    \includegraphics[width=1\linewidth]{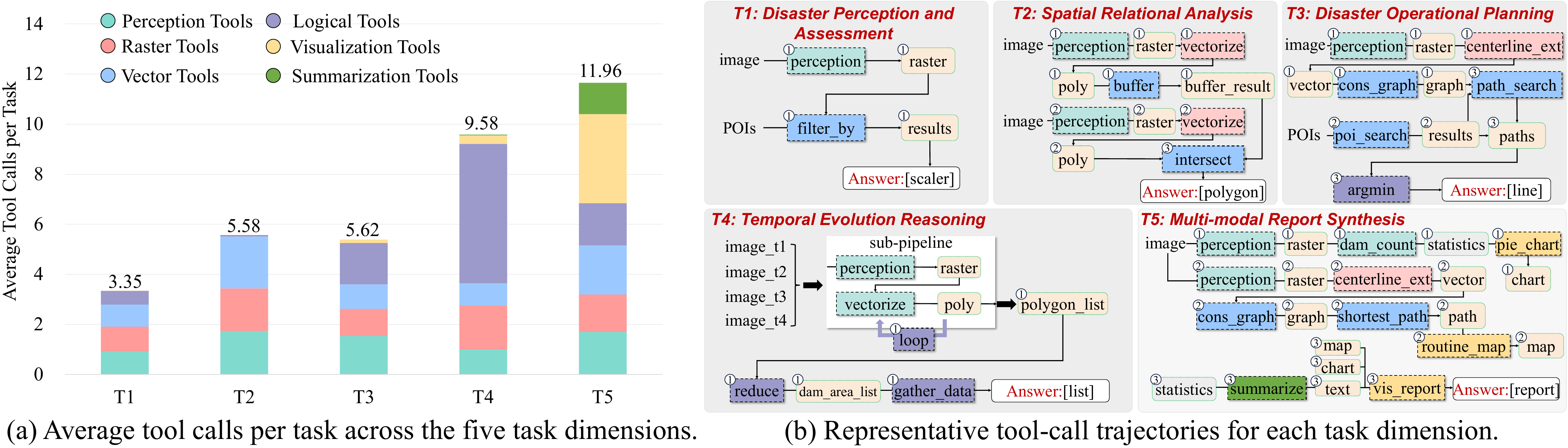}
    \caption{Complexity and tool-usage profiles across DORA's five dimensions. (a)~Trajectory length grows from $3.35$ (T1) to $11.96$ steps (T5), with distinct tool-category distributions per dimension.
(b)~Representative trajectories show the progression from linear perception chains (T1) to full-stack report synthesis (T5); node colors match (a).}
\label{fig:tool_distr_traj}
\end{figure*}

\noindent \textbf{Analytical dimensions.}
Fig.~\ref{fig:analytical_dim} classifies the 515 tasks into five task dimensions, each with a distinct tool-usage profile, reasoning pattern, and output modality. This taxonomy is motivated by two rationales.
\emph{Substantively}, the five dimensions mirror canonical stages of geospatial information processing (damage perception, spatial modeling, decision support, temporal analysis, and integrated reporting), each recognized as a distinct operational need by international disaster-response frameworks (UNOSAT~\cite{unosat}, Copernicus EMS~\cite{copernicus_ems}, FEMA ICS~\cite{fema_usr}, UN OCHA~\cite{ocha}).
\emph{Methodologically}, each dimension isolates a progressively more demanding agent capability: atomic tool invocation and output parsing (T1); tool composition with spatial semantics (T2); operational knowledge grounding (T3); temporal abstraction and iterative state tracking (T4); and cross-modal synthesis with structured generation (T5). This capability hierarchy manifests as increasing trajectory length and broadening tool-category coverage across T1--T5 (Fig.~\ref{fig:tool_distr_traj}(a)).

\noindent \textbf{Task Taxonomy.}
Fig.~\ref{fig:tool_distr_traj} (b) shows some representative trajectories for each dimension, respectively.

\noindent \emph{T1: Disaster Perception \& Assessment (PA).} T1 covers atomic geospatial quantification from single- or bi-temporal imagery. Typical tasks extract damaged areas, building counts, or debris volumes via \texttt{perception} and \texttt{raster} tools. More complex instances cross-reference perception outputs with POI records to assess damage at specific facilities (e.g., ``how many hospitals fall within the severe-damage zone''), following UNOSAT rapid structural damage grading products~\cite{unosat}.

\noindent \emph{T2: Spatial Relational Analysis (SR).} T2 requires composing multiple perception and GIS tools across data layers with explicit spatial semantics such as proximity, containment, and overlay, following the multi-layer exposure analysis in FEMA's Hazus framework~\cite{fema_hazus}. A representative pattern first segments two thematic layers independently (e.g., lava extent and building footprints), vectorizes each, then applies buffer and intersection operations to answer queries like ``identify intact buildings that lie within 100 m of the lava flow boundary.''

\noindent \emph{T3: Disaster Operational Planning (OP).} T3 translates geospatial outputs into actionable resource allocation, logistics, or route planning decisions, invoking \texttt{logical} tools and probing operational knowledge under domain-specific constraints (clearance rates, vehicle capacities, shortest accessible routes), following following FEMA's Urban Search and Rescue System~\cite{fema_usr}.

\noindent \emph{T4: Temporal Evolution Reasoning (TE).} T4 tracks disaster dynamics across 3--5 observation phases (e.g., flood progression, wildfire spread, or post-disaster reconstruction) by iteratively invoking the same analytical sub-pipeline per phase and aggregating cross-phase results to identify trends, peaks, and phase transitions, following Copernicus EMS Monitoring products~\cite{copernicus_ems}.

\noindent \emph{T5: Multi-modal Report Synthesis (RS).} T5 integrates capabilities from T1--T4 and additionally requires \texttt{visualization} and \texttt{summarization} tools to produce situation reports with damage maps, trend charts, and narrative summaries, following UN OCHA reporting standards~\cite{ocha}. T5 yields the longest average trajectories in the benchmark.

\subsection{Tool library}
\label{sec:tools}
DORA provides a library of 108 MCP-compliant tools spanning six functional categories (Table~\ref{tab:tool_library}): 1)~\texttt{Perception}, semantic segmentation models for buildings, roads, floods, landslides, lava, vehicles, and other disaster-relevant objects from optical, SAR, or multi-sensor imagery. These segmentation models were trained on public datasets (xBD~\cite{gupta2019xbd}, DisasterM3~\cite{wang2025disasterm3}, BRIGHT~\cite{chen2025bright}, GVLM~\cite{ZHANG20231}, RescueNet~\cite{rahnemoonfar2023rescuenet}, etc) or borrowed from the challenge champion solutions (Landslide4Sense~\cite{ghorbanzadeh2022outcome}, SpaceNet-7~\cite{vanetten2021spacenet7}).
\textbf{To prevent data leakage,} we train all perception tools only on the public \emph{training} splits and construct DORA tasks \emph{exclusively} from the held-out \emph{test} splits.
2)~\texttt{Raster}: pixel-level operations including area, zonal statistics, thresholding, grid windowing, and vectorization. 3)~\texttt{Vector}: geometric operations such as buffering, intersection, connected-component analysis, POI querying, and graph-based path routing. 4)~\texttt{Logical}: control-flow primitives (logi.loop, logi.reduce, logi.truck\_trips) that enable iterative reasoning and domain-specific planning. 5)~\texttt{Visualization}: map rendering, charting, and multi-panel report layout. 6)~\texttt{Summarization}: fixed report-generation tools for evidence extraction and narrative rendering, shared by all agents as part of the environment.
All tools are implemented as MCP servers, exposing a uniform JSON-RPC interface with typed input and output schemas. 
This design decouples tool implementation from agent logic, allowing any MCP-compatible agent framework to be evaluated without modification.

\begin{table}[h]
\centering
\small
\caption{Overview of the DORA tool library.}
\label{tab:tool_library}
\resizebox{1.0\linewidth}{!}{
\begin{tabular}{llllll}
\toprule
\textbf{Category} & \textbf{\#} & \textbf{Scope} & \textbf{Representative Tools} & \textbf{Output} & \textbf{Implementation} \\
\midrule
\texttt{Perception} & 31 & \makecell[l]{Semantic segmentation of \\ disaster-relevant objects} &  \makecell[l]{seg.building\_damage, seg.flood,\\ seg.road\_damage} & Raster mask & \makecell[l]{DinoV3, SegFormer,\\ HRNet, SwinUperNet}\\ \midrule
\texttt{Raster} & 18 & \makecell[l]{Raster algebra and conversion} & ras.area, ras.diff, ras.vectorize & scalar, polygon & GDAL, Rasterio \\ \midrule
\texttt{Vector} & 31 & \makecell[l]{GIS operations, graph construction,\\ POI querying} &  \makecell[l]{vec.intersect, vec.shortest\_path,\\ poi.filter\_by\_damage} & \makecell[l]{scalar, point, \\line, polygon} & \makecell[l]{Shapely, NetworkX,\\ GeoPandas} \\ \midrule
\texttt{Logical} & 15 & Control flow and planning & logi.loop, logi.reduce & decision & Python \\ \midrule
\texttt{Visualization} & 11 & Map rendering and report layout &  \makecell[l]{vis.damage\_map, vis.route\_map, \\vis.report\_page} & image & Matplotlib \\ \midrule
\texttt{Summarization} & 2 & \makecell[l]{Evidence extraction and rendering} & \makecell[l]{m.extract\_evidence,\\ m.summarize} & text & Fixed report backend \\
\bottomrule
\end{tabular}}

\end{table}

\begin{figure*}[b]
    \centering
    \includegraphics[width=1\linewidth]{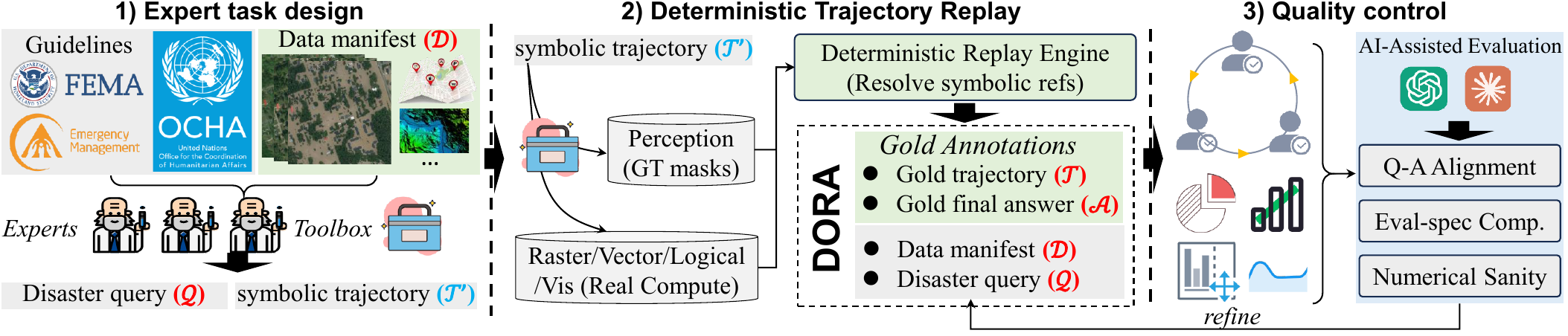}
    \caption{Our annotation pipeline: (1)~experts author queries and symbolic trajectories; (2)~a deterministic replay engine resolves references to produce gold annotations; (3)~AI-assisted evaluation.}
\label{fig:annotation_check}
\end{figure*}

\subsection{Annotation Pipeline}
Building a reliable tool-use benchmark requires ground-truth trajectories that are both \emph{semantically faithful} (each tool call reflects a genuine analytical step) and \emph{numerically grounded} (observation values come from real ground truths).
DORA's construction pipeline achieves this through a three-stage process illustrated in Fig.~\ref{fig:annotation_check}.
\textbf{1) Expert task design.} Each task is authored by domain experts with remote-sensing and disaster-management backgrounds. Given a disaster scene and data manifest $\mathcal{D}$, the expert writes: (i)~a natural-language query $\mathcal{Q}$ grounded in an operational need; (ii)~a gold tool-call sequence $\mathcal{T}^{*} = \langle t_1, \dots, t_K \rangle$ specifying each tool name, its purpose, and its input arguments. When an argument depends on a previous tool's output, it is written as a symbolic reference (e.g., \texttt{<mask\_path> from trajectory[2].obs}). This template-based design keeps the trajectory logically complete and decoupled from any execution environment.
\textbf{2) Fill-the-blank execution.}
A deterministic replay engine resolves all symbolic references and executes the trajectory. 
During annotation construction, perception calls are replaced by human-annotated GT-mask lookups, while all downstream tools run normally to produce the curated reference answer $\mathcal{A}$. This GT-mask replay is used only for annotation and is never exposed to agents during evaluation.
\textbf{3) Quality control.}
We perform systematic cross-validation over question-answer alignment, eval-spec completeness (zero \texttt{unknown}-type fields), and numerical sanity checks.

\section{Experiments}
\label{sec:experiments}
\noindent \textbf{Evaluation Protocol.}
Following prior work~\cite{ma2024m,wang2024gta,qintoolllm,feng2026earthagent}, we adopt a dual-level protocol covering both reasoning trajectory and final answer.
\textbf{1) Trajectory metrics.}
Given gold $\mathcal{T}^\star$ and predicted $\mathcal{T}^\text{pred}$ trajectories, we report four measures~\cite{feng2026earthagent}: Tool-Any-Order (order-agnostic tool-set recall), Tool-In-Order (longest common subsequence of tool identifiers, normalized by $|\mathcal{T}^\star|$), Tool-Exact-Match (longest matching prefix length, normalized by $|\mathcal{T}^\star|$), and Parameter Accuracy (per-step argument matching, conditioned on correct tool names).
\textbf{2) Final answer metrics.}
The seven typed fields of $\mathcal{A}^{*}$ defined in \S\ref{sec3:tasks} reduce to four atomic scoring operators (Tab.~\ref{tab:eval_atoms}): \emph{scalar} closeness, normalized \emph{string} exact-match, \emph{point} Euclidean-distance gate, and \emph{polygon} IoU gate. Composite types inherit these atoms: a \emph{dict} averages per-key scalar scores over the key union; a \emph{set} is scored by F$_1$ over string-matched elements; a \emph{line} decomposes into start point, end point, and total length. For free-form $T_5$ summaries, we extract key statistics and categorical labels for deterministic scoring. Since multiple trajectories can be valid, we treat final-answer metrics as primary and trajectory metrics as diagnostic.
\textbf{3) Efficiency.}
$\textsc{Eff} = |\mathcal{T}^\star| / \max(|\mathcal{T}^\text{pred}|, |\mathcal{T}^\star|) \in (0,1]$ rewards agents that reach correct answers without redundant tool calls. 

\noindent \textbf{Evaluation Methods}
We evaluate 13 LLMs spanning commercial models (GPT-5.4 series~\cite{openai2026gpt54}, Claude-Sonnet-4.6~\cite{anthropic2026claude46sonnet}, Gemini-3.0-Flash~\cite{google2026gemini3flash}, Grok-4.1 Fast~\cite{xai2025grok41}) and
open-source models (Qwen3.5-series~\cite{qwen2026qwen35}, MiMo-V2-Pro~\cite{xiaomi2026mimov2pro}, Step-3.5-Flash~\cite{stepfun2025step35flash}, DeepSeek-V3.2~\cite{deepseek2025v32}, Gemma-4-31B~\cite{gemma2026gemma4}, GPT-OSS-120B~\cite{openai2025gptoss}, MiniMax-M2.7~\cite{minimax2026m27}).
Two advanced vision-language models (Qwen3-VL-235B~\cite{qwen2025qwen3vl} and Gemini-3.0~Flash~\cite{google2026gemini3flash}) receive the raw imagery and question without access to any tools, serving as a tool-free baseline on visual reasoning.
We also report Gold Trajectory that executes the expert-authored tool sequence. It uses model-backed perception tools rather than GT masks, making it a planning-and-argument oracle rather than a perfect-answer oracle.
All agents are implemented under a ReAct-style~\cite{yao2023react} agent loop. In addition, we evaluate three alternative scaffolds (Plan-then-Execute~\cite{he2025plan}, Reflexion~\cite{shinn2023reflexion}, and ReWOO~\cite{xu2023rewoo}) to test mainstream agent paradigms.

\begin{table}[t]
\centering
\small
\caption{Atomic scoring operators used by DORA. All typed fields are ultimately reduced to these four rules. $y_p, y_g$ denote predicted and gold values; $\mathrm{nrm}(\cdot)$ applies string normalization.}
\label{tab:eval_atoms}
\resizebox{0.7\linewidth}{!}{
\begin{tabular}{@{}l l l@{}}
\toprule
\textbf{Operator} & \textbf{Scoring function} & \textbf{Parameter} \\
\midrule
scalar &
$|y_p - y_g| \le \tau_r\,|y_g|$ &
$\tau_r{=}0.2$ \\[3pt]
string &
$\mathrm{nrm}(y_p) = \mathrm{nrm}(y_g)$ &
-- \\[3pt]
point &
$\|(y_p^x, y_p^y) - (y_g^x, y_g^y)\|_2 \le \tau_{\text{px}}\,$ &
$\tau_{\text{px}}{=}20\,\text{px}$ \\[3pt]
polygon &
$\mathrm{IoU}(y_p, y_g) \ge \tau_{\text{IoU}}\,$ &
$\tau_{\text{IoU}}{=}0.5$ \\[3pt]
composite &
mean of atomic scores over expanded elements &
inherits atoms \\
\bottomrule
\end{tabular}}
\end{table}

\begin{table*}[h]
\centering
\caption{Main results on the DORA benchmark. We report final-answer accuracy (\%) per task dimension, trajectory metrics (\%), and efficiency.}
\label{tab:main}
\resizebox{1.0\linewidth}{!}{
\begin{tabular}{l|c|ccccc|ccccc}
\toprule
& \multicolumn{6}{c|}{\textbf{Final Answer (\%)}} & \multicolumn{5}{c}{\textbf{Trajectory (\%)}} \\
\cmidrule(lr){2-7} \cmidrule(lr){8-12}
\textbf{Model} & AVG & $T_1$(PA) & $T_2$(SR) & $T_3$(OP) & $T_4$(TE) & $T_5$(RS) & T-Any-O & T-In-Ord & T-Exact-M & ParAcc & Eff. \\
\midrule
\rsb Baselines &&&&&&&&&&& \\
Gold Trajectory  & 80.48 & 71.31 & 63.19 & 83.63 & 90.77 & 93.50 & 100   & 100   & 100   & 100   & 100   \\
Gemini-3.0-Flash~\cite{google2026gemini3flash}        & 18.55 & 5.98  & 19.91 & 17.48 & 29.36 & 20.03 & \multicolumn{5}{c}{\emph{Single-step prediction without tool use.}} \\
Qwen3-VL-235B~\cite{qwen2025qwen3vl}                  & 18.30 & 5.53  & 28.63 & 12.63 & 27.75 & 16.95 & \multicolumn{5}{c}{\emph{Single-step prediction without tool use.}} \\
\midrule
\gb Commercial Models &&&&&&&&&&& \\
Gemini-3.0-Flash~\cite{google2026gemini3flash}        & \textbf{53.74} & \underline{54.19} & 49.00 & \textbf{59.82} & 60.40 & 45.31 & 75.17 & 66.57 & \textbf{35.55} & \textbf{42.63} & 83.05 \\
Grok-4.1-Fast~\cite{xai2025grok41}                    & 52.10 & 53.07 & \textbf{53.40} & 54.23 & 55.28 & 44.52 & 59.15 & 53.19 & 28.09 & 31.96 & 86.83 \\
GPT-5.4~\cite{openai2026gpt54}                        & 47.63 & 52.85 & \underline{50.80} & 53.50 & 51.90 & 29.11 & 66.84 & 59.43 & 31.82 & 37.71 & \underline{88.46} \\
GPT-5.4-Nano~\cite{openai2026gpt54}                   & 38.14 & 44.40 & 33.41 & 39.98 & 45.55 & 27.37 & 54.96 & 49.06 & 21.81 & 27.02 & 84.36 \\
Claude-Sonnet-4.6~\cite{anthropic2026claude46sonnet}  & 52.01 & \textbf{54.43} & 48.23 & 51.82 & 60.05 & \underline{45.53} & \textbf{76.29} & 66.92 & 32.28 & \underline{39.48} & 75.32 \\
\midrule
\ftb Open-Source Models &&&&&&&&&&& \\
Qwen3.5-397B-A17B~\cite{qwen2026qwen35}               & \underline{53.45} & 51.03 & \textbf{53.40} & 51.82 & \underline{61.83} & \textbf{49.17} & \underline{75.44} & \underline{66.97} & 32.13 & 36.50 & 76.54 \\
Qwen3.5-35B-A3B~\cite{qwen2026qwen35}                 & 24.01 & 14.15 & 13.89 & 33.08 & 37.62 & 21.30 & 55.20 & 48.91 & 23.10 & 27.08 & 80.39 \\
Gemma-4-31B~\cite{gemma2026gemma4}                    & 51.17 & 51.46 & 49.00 & 52.33 & 61.03 & 42.03 & 69.89 & 62.58 & 32.82 & 38.23 & 86.43 \\
MiMo-V2-Pro~\cite{xiaomi2026mimov2pro}                & 52.89 & 53.68 & 47.43 & \underline{55.48} & \textbf{63.58} & 44.26 & 74.05 & \textbf{67.13} & \underline{34.24} & 38.45 & 79.11 \\
MiniMax-M2.7~\cite{minimax2026m27}                    & 48.35 & 51.87 & 48.53 & 50.16 & 50.59 & 40.62 & 62.89 & 55.19 & 25.32 & 29.68 & 79.33 \\
DeepSeek-V3.2~\cite{deepseek2025v32}                  & 48.23 & 49.80 & 49.15 & 50.57 & 50.62 & 41.01 & 74.26 & 65.99 & 30.39 & 34.69 & 72.44 \\
Step-3.5-Flash~\cite{stepfun2025step35flash}          & 46.68 & 49.70 & 44.91 & 48.33 & 48.90 & 41.58 & 62.59 & 55.91 & 25.33 & 29.43 & 78.29 \\
GPT-OSS-120B~\cite{openai2025gptoss}                  & 35.11 & 42.70 & 37.58 & 42.00 & 24.43 & 28.84 & 48.69 & 43.60 & 22.79 & 26.24 & \textbf{90.42} \\
\bottomrule
\end{tabular}}
\end{table*}

\subsection{Benchmark results}
\noindent \textbf{DORA is challenging for all models.}
As shown in Tab.~\ref{tab:main},  
even the strongest model, Gemini-3.0-Flash, achieves only 53.74\% average accuracy, $~$26\% below the gold trajectory. The tool-free baselines Qwen3-VL-235B and Gemini-3.0-Flash without tools confirm that visual reasoning alone cannot substitute for compositional tool use. GPT-OSS-120B attains the highest efficiency with a low accuracy, indicating that agents confidently select few tools but often the \emph{wrong} ones. On the open-source side, Qwen3.5-397B-A17B narrows the gap to Gemini to within 0.3\%, Gemma-4-31B and MiMo-V2-Pro form a strong second tier, and all three surpass GPT-5.4-Nano by more than 12\%; strikingly, the 31B Gemma also outperforms GPT-OSS-120B despite using 4$\times$ fewer parameters, indicating that agentic post-training quality matters more than raw parameter scale.

\noindent \textbf{Tool heterogeneity amplifies compositional fragility.}
Across all models, \emph{Tool-Any-Order} exceeds \emph{Tool-Exact-Match} by over 30\% on average, indicating that even when agents know which tools to call, they rarely get the order and arguments right. \emph{Parameter Accuracy} reveals the same fragility: argument grounding (damage indices, GSD, path references, etc.) remains unreliable. This fragility compounds with \emph{compositional diversity} rather than raw length alone. Although $T_4$ involves long trajectories, its steps largely consist of repeated invocations of the same tools across multiple temporal phases, a simple and regular pattern that agents handle comparatively well. In contrast, $T_5$ chains \emph{distinct} tools spanning perception, analysis, visualization, and summarization, where a single early error cascades through a heterogeneous pipeline. All models suffer large drop on $T_5$, confirming that \emph{tool heterogeneity} is the dominant source of compositional failure.

\begin{figure*}[hbt]
    \centering
    \begin{minipage}{0.50\linewidth}
        \centering
        \includegraphics[width=\linewidth]{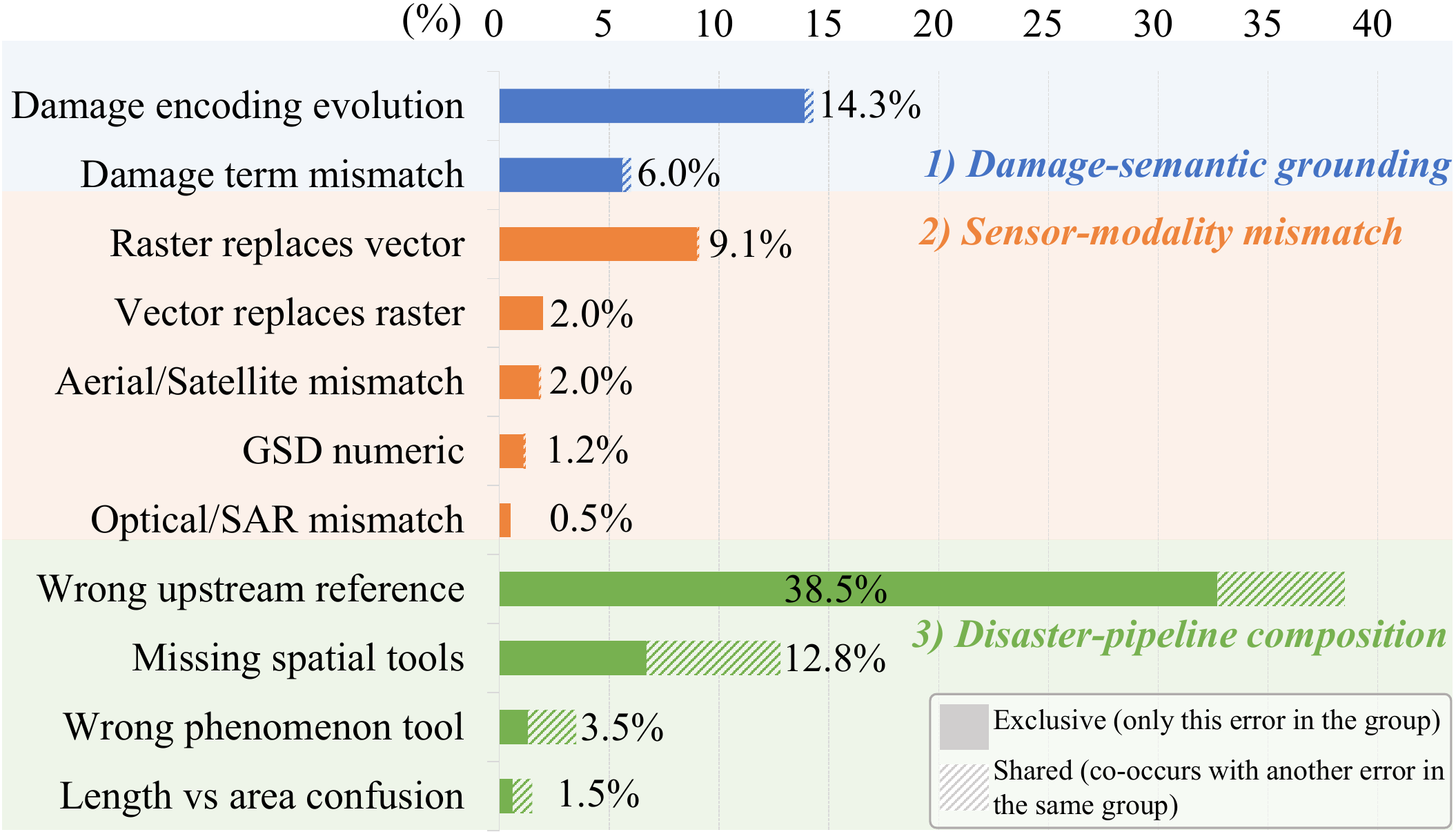}
        \caption{Failure modes in disaster domain.}
        \label{fig:failure_mode}
    \end{minipage}
    \hfill
    \begin{minipage}{0.45\linewidth}
        \centering
        \refstepcounter{table}\label{tab:ap_if}
        {\normalsize\centering Table~\thetable. AP versus IF mode.\par}
        \vspace{6pt}
        \resizebox{\linewidth}{!}{%
       \begin{tabular}{l|ccccc}
\toprule
\textbf{Protocol} & T-Any-O & T-In-Ord & T-Exact-M & ParAcc & AVG \\
\midrule
Gold Trajectory  & 100 & 100 & 100 & 100 & 80.48 \\ \midrule
\textit{Gemini-3.0-Flash}& & & && \\
\quad w.\ AP  & 75.17 & 66.57 & 35.55 & 42.63 & 53.74 \\
\quad w.\ IF  & 82.56 & 82.03 & 62.53 & 53.49 & 55.71 \\
\quad $\Delta_{\text{AP}\to\text{IF}}$ & \textcolor{red}{$\uparrow$7.39} & \textcolor{red}{$\uparrow$15.46} & \textcolor{red}{$\uparrow$26.98} & \textcolor{red}{$\uparrow$10.86} & \textcolor{red}{$\uparrow$1.97} \\
\midrule
\textit{Gemma-4-31B} & & & &&\\
\quad w.\ AP  & 69.89 & 62.58 & 32.82 & 38.23 & 51.17 \\
\quad w.\ IF  & 79.80 & 79.63 & 61.76 & 50.24 & 52.25 \\
\quad $\Delta_{\text{AP}\to\text{IF}}$ & \textcolor{red}{$\uparrow$9.91} & \textcolor{red}{$\uparrow$17.05} & \textcolor{red}{$\uparrow$28.94} & \textcolor{red}{$\uparrow$12.01} & \textcolor{red}{$\uparrow$1.08} \\
\midrule
\textit{MiniMax-M2.7}& & & && \\
\quad w.\ AP  & 62.89 & 55.19 & 25.32 & 29.68 & 48.35 \\
\quad w.\ IF  & 80.85 & 79.83 & 50.95 & 43.98 & 52.75 \\
\quad $\Delta_{\text{AP}\to\text{IF}}$ & \textcolor{red}{$\uparrow$17.96} & \textcolor{red}{$\uparrow$24.64} & \textcolor{red}{$\uparrow$25.63} & \textcolor{red}{$\uparrow$14.30} & \textcolor{red}{$\uparrow$4.40} \\
\bottomrule
\end{tabular}}
\end{minipage}
\end{figure*}

\noindent \textbf{Disaster-domain grounding exposes unique failure modes.}
Beyond generic compositional fragility, DORA exposes three disaster-domain failure modes~(Fig.~\ref{fig:failure_mode}): 1)~\emph{Damage-semantic grounding} (20.3\%): agents struggle to map damage terminology (``partial damaged'', ``total destroyed'', ``affected'', ``flooded'') to class-index lists, and how indices evolve as multi-class masks $\{1, 2, 3\}$ are binarized $\{0, 255\}$ by tools like \texttt{ras.compose\_classes} or \texttt{ras.threshold}; 14.3\% pass stale indices to vectorization while 6.0\% select wrong granularity.
2)~\emph{Sensor-modality mismatch} (14.8\%): 9.1\% mistake raster for vector relational operations, while 5.7\% involve modality swaps (optical$\leftrightarrow$SAR, aerial$\leftrightarrow$satellite, bi-temporal$\leftrightarrow$multi-spectral) or GSD mis-specification.
3)~\emph{disaster-pipeline composition} (56.3\%): agents mis-decompose compound spatial concepts, 12.8\% miss key tools \texttt{vec.intersect} and \texttt{vec.buffer}; 38.5\% link downstream tools to the wrong upstream mask; and 3.5\% pick the wrong phenomenon tool 
(e.g., invoking \texttt{seg.landslide} on a hurricane scene without any landslides). 
These show that LLM agents lack disaster-semantic grounding for operational reasoning.

\subsection{Ablation Study}
\noindent \textbf{Even with the right tools, argument grounding remains highly challenging.}
We evaluate agents under two protocols: \emph{auto-planning} (AP, the default) and \emph{instruction-following} (IF, additionally given only the gold tool-order hints and must still issue tool calls and infer all arguments themselves). Tab.~\ref{tab:ap_if} reveals a striking decoupling: IF yields substantial trajectory gains, confirming that agents can follow a given pipeline once tools are specified. Yet final-answer accuracy improves by only 1.08--4.40\%, leaving all three models 25--28\% below the gold trajectory. Even after tool-order uncertainty is reduced, inferring correct arguments (file paths, damage class indices, GSD parameters) and extracting the right intermediate outputs remain dominant residual failure modes. DORA is therefore doubly bottlenecked: agents must both select the right tools \emph{and} use the right arguments, with errors at either step propagating downstream.

\begin{figure*}[htb]
    \centering
    \includegraphics[width=1\linewidth]{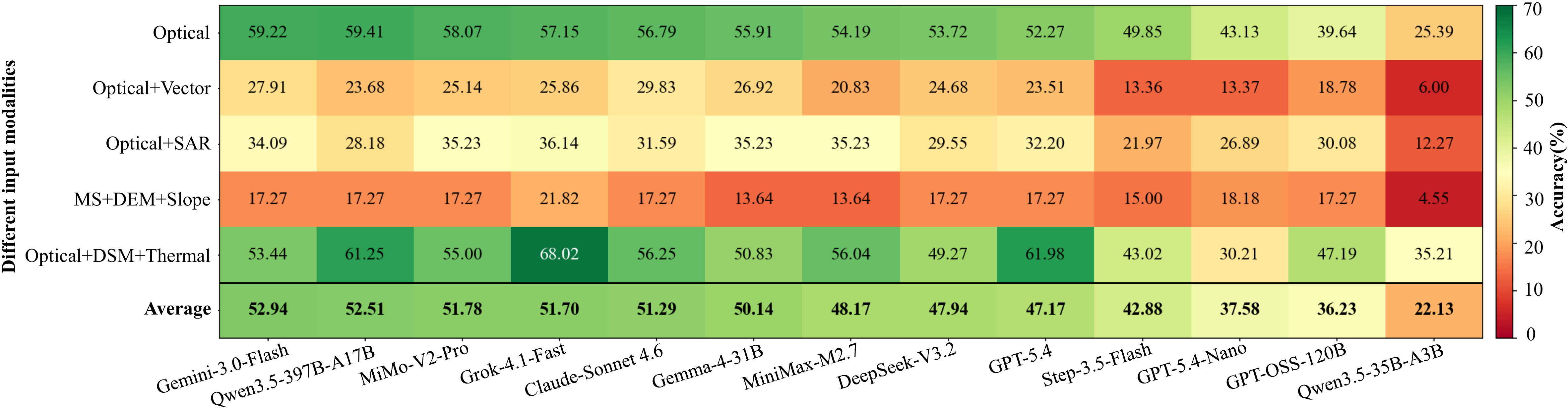}
    \caption{Agent performances (\%) decomposed by input modality configuration.}
\label{fig:modality_config}
\end{figure*}
\noindent\textbf{Accuracy varies sharply with modality configuration.}
Fig.~\ref{fig:modality_config} decomposes agent performance by sensor configuration. Auxiliary modalities universally degrade accuracy: the pure-optical baseline reaches 52--59\%, but adding OSM vector layers drops accuracy to 18--27\%, and optical-SAR fusion to 24--36\%. Additional modalities demand longer planning over a broader tool space, amplifying grounding errors at each step. All agents struggle with MS+DEM+Slope: these tasks chain geometric operations like computing susceptibility over terrain strata and composing landslide pipelines beyond image perception. Thermal-augmented tasks remain tractable because land-surface temperature has clear numerical semantics ($^\circ$C) aligned with generic knowledge priors, reducing specialized grounding needs. Current agents handle pure-optical perception well but struggle with symbolic fusion (OSM), cross-sensor alignment (SAR), and compositional geometric reasoning (slope, DEM).

\begin{figure*}[hbt]
    \centering
    \begin{minipage}{0.45\linewidth}
        \centering
        \includegraphics[width=\linewidth]{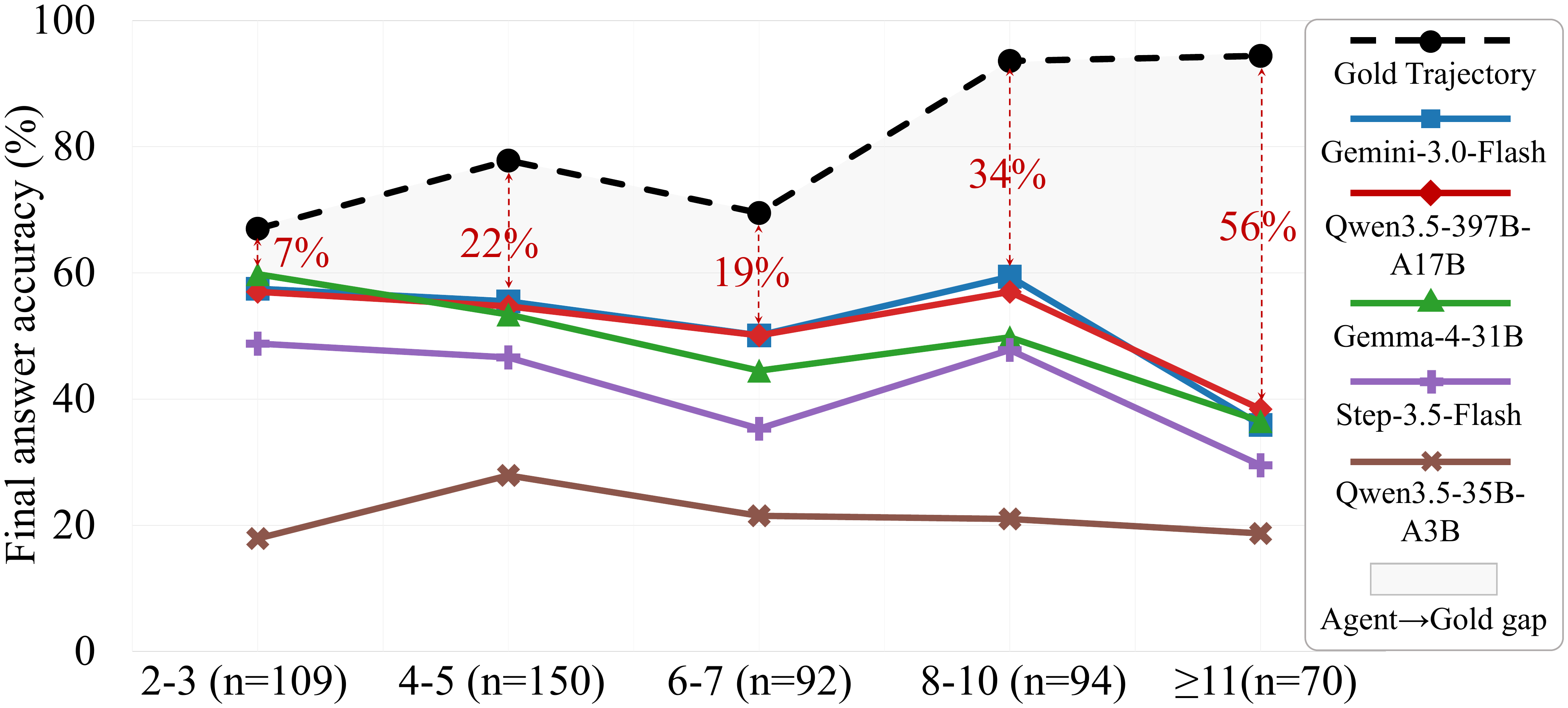}
        \caption{Agents vs.\ gold trajectory: compositional fragility scales with length.}
        \label{fig:traj_length}
    \end{minipage}
    \hfill
    \begin{minipage}{0.49\linewidth}
        \centering
        \refstepcounter{table}\label{tab:scaffold_ablation}
        {\normalsize\centering Table~\thetable. Agent scaffolding ablation.\par}
        \vspace{6pt}
        \resizebox{\linewidth}{!}{%
       \begin{tabular}{l|cccc}
\toprule
\textbf{Scaffold} & AVG & T-Exact-M & ParAcc & Latency (s/task) \\
\midrule
ReAct  & 51.17 & 32.82 & 38.23 & 80 \\ \midrule
PE~\cite{he2025plan} & \textbf{54.41} & \textbf{38.83} & \textbf{47.27} & 179 \\
$\Delta$ & \textcolor{red}{$\uparrow$3.24} & \textcolor{red}{$\uparrow$6.01} & \textcolor{red}{$\uparrow$9.04} & \textcolor{blue}{$\uparrow$99} \\
\midrule
Reflexion~\cite{shinn2023reflexion} & 52.74 & 36.21 & 40.05 & 167 \\
$\Delta$ & \textcolor{red}{$\uparrow$1.57} & \textcolor{red}{$\uparrow$3.39} & \textcolor{red}{$\uparrow$1.82} & \textcolor{blue}{$\uparrow$87} \\
\midrule
ReWOO~\cite{xu2023rewoo} & 44.26 & 25.18 & 25.92 & 82 \\
$\Delta$ & \textcolor{blue}{$\downarrow$6.91} & \textcolor{blue}{$\downarrow$7.64} & \textcolor{blue}{$\downarrow$12.31} & \textcolor{blue}{$\uparrow$2} \\
\bottomrule
\end{tabular}%
        }
    \end{minipage}
\end{figure*}

\noindent \textbf{Compositional fragility scales with trajectory length.}
Fig.~\ref{fig:traj_length} bins all tasks by gold-trajectory length and reports per-bucket final-answer accuracy across representative models. The agent-to-gold gap widens dramatically with length: top models track the gold ceiling within 7\% on short pipelines (2--3 steps) but fall 56\% behind at $\geq$11 steps. Notably, the gold ceiling itself \emph{rises} on long trajectories, so the widening gap reflects agent-side compositional decay rather than harder underlying tasks: each additional tool call multiplies the chance of an upstream argument or sequencing error propagating to the final answer. 
Hence, the 8--10 and $\geq$11 buckets exhibit the largest gaps, confirming long-horizon synthesis as the most fragile compositional regime.

\noindent \textbf{Scaffolding helps modestly but does not close the gap.}
We evaluate three scaffolds on Gemma-4-31B, i.e., Plan-then-Execute (PE)~\cite{he2025plan}, Reflexion~\cite{shinn2023reflexion} and ReWOO~\cite{xu2023rewoo} (Tab.~\ref{tab:scaffold_ablation}).
PE provides the largest gain (+3.24\%), as upfront planning avoids agent misdirection from intermediate observations on DORA's long compositional pipelines.
Reflexion offers a marginal +1.57\% at 2.1$\times$ latency, as self-correction yields limited returns when failures stem from disaster-domain grounding rather than reasoning slips.
ReWOO collapses, confirming that removing observation feedback hurts data-heavy tasks driven by intermediate masks and geometries.
Critically, even the best scaffold trails the gold ceiling by over 26\%, showing the dominant bottleneck remains domain-specific knowledge and tool-argument grounding, not scaffolding strategy.

\section{Conclusion}
We presented DORA, the first agentic benchmark for operational disaster response, with 515 expert-authored tasks, 108 disaster-tailored tools, and five analytical dimensions over heterogeneous geospatial data. Evaluating 13 frontier LLMs reveals three persistent challenges: disaster-domain grounding exposes unique failure modes; agents are doubly bottlenecked by tool selection and argument usage; and compositional fragility scales sharply with trajectory length, with the gold-ceiling gap reaching 56\% on long pipelines. We will release DORA with all data, tools, and evaluation protocols to drive progress toward operationally reliable disaster-response AI agents.

\section*{Acknowledgments}
This work was supported by JST CRONOS (Grant Number JPMJCS25K5), JST NEXUS (Grant Number JPMJNX25CA), and KAKENHI (25K03145, 26K21244). Weihao Xuan is supported by RIKEN Junior Research Associate (JRA) Program. Pengyu Dai is supported by RIKEN Incentive Research Project 2026. We also thank Ritwik Gupta for sharing the valuable xBD dataset and for his
expertise in disaster response guidance.

\bibliographystyle{ieeetr}
\bibliography{references}

\end{document}